\documentclass[pdflatex,sn-mathphys-num]{sn-jnl}

\usepackage{graphicx}%
\usepackage{multirow}%
\usepackage{amsmath,amssymb,amsfonts}%
\usepackage{amsthm}%
\usepackage{mathrsfs}%
\usepackage[title]{appendix}%
\usepackage{xcolor}%
\usepackage{textcomp}%
\usepackage{manyfoot}%
\usepackage{booktabs}%
\usepackage{algorithm}%
\usepackage{algorithmicx}%
\usepackage{algpseudocode}%
\usepackage{listings}%

\theoremstyle{thmstyleone}%
\theoremstyle{thmstyletwo}%

\theoremstyle{thmstylethree}%

\begin{document}

\title[Article Title]{Recognizing Pneumonia in Real-World Chest X-rays with a Classifier Trained with Images Synthetically Generated by Nano Banana}


\author[1,2]{\fnm{Jiachuan} \sur{Peng}}\email{Jiachuan.Peng@mbzuai.ac.ae}

\author[3]{\fnm{Kyle} \sur{Lam}}\email{k.lam@imperial.ac.uk}

\author[1]{\fnm{Jianing} \sur{Qiu}}\email{Jianing.Qiu@mbzuai.ac.ae}

\affil[1]{\orgname{Mohamed bin Zayed University of Artificial Intelligence}, \orgaddress{\country{UAE}}}

\affil[2]{\orgname{University of Oxford}, \orgaddress{\country{UK}}}

\affil[3]{\orgname{Imperial College London}, \orgaddress{\country{UK}}}


\abstract{We trained a classifier with synthetic chest X-ray (CXR) images generated by Nano Banana, the latest AI model for image generation and editing, released by Google. When directly applied to real-world CXRs having only been trained with synthetic data, the classifier achieved an AUROC of 0.923 (95\% CI: 0.919 - 0.927), and an AUPR of 0.900 (95\% CI: 0.894 - 0.907) in recognizing pneumonia in the 2018 RSNA Pneumonia Detection dataset (14,863 CXRs), and an AUROC of 0.824 (95\% CI: 0.810 - 0.836), and an AUPR of 0.913 (95\% CI: 0.904 - 0.922) in the Chest X-Ray dataset (5,856 CXRs). These external validation results on real-world data demonstrate the feasibility of this approach and suggest potential for synthetic data in medical AI development. Nonetheless, several limitations remain at present, including challenges in prompt design for controlling the diversity of synthetic CXR data and the requirement for post-processing to ensure alignment with real-world data. However, the growing sophistication and accessibility of medical intelligence will necessitate substantial validation, regulatory approval, and ethical oversight prior to clinical translation.}

\keywords{Pneumonia Recognition, Text-to-Image Generation, Generative AI}

\maketitle

\section{Introduction}\label{sec1}

Recent advances in large vision-language models (VLMs) have substantially improved the realism and clinical fidelity of text-to-image generation. Trained on billions of image-text pairs, these models now have the capacity to synthesize highly realistic medical images including chest X-rays (CXRs)~\cite{xie2025sana}. In the medical domain, Synthetic image generation is an emerging area within clinical research, supporting data augmentation and facilitating cross-center data sharing while mitigating privacy concerns~\cite{bluethgen2025vision}. 

Early work has shown that latent diffusion models can synthesize CXRs that, when conditioned on radiology reports, achieve a level of image quality and clinical coherence~\cite{bluethgen2025vision}. Similar progress has been made in ophthalmology, where synthetic eye images have achieved diagnostic performance comparable to real images. Medical foundation models, such as VisionFM, have demonstrated that judicious integration of synthetic data with real data can enhance model utility and robustness across diverse clinical applications~\cite{qiu2024development}.

The emergence of increasingly large-scale commercial generative models trained on diverse internet datasets has created a novel opportunity where medical image-text pairs are now embedded within these foundation models. As a result, off-the-shelf AI tools, such as Nano Banana, can generate meaningful medical images, including CXRs, without requiring domain-specific fine-tuning. This capability mirrors earlier successes in leveraging Internet-sourced datasets, such as Medical Twitter, to build pathology foundation models ~\cite{huang2023visual}, demonstrating a precedent for democratizing medical resources through heterogeneous data sources.

The widespread accessibility of such generative capabilities has significant implications. Synthetic medical data can now be potentially obtained directly by clinicians and researchers without requiring specialized technical expertise, therefore democratizing access to high-quality training and validation datasets and enabling non-technical users to advance scientific progress in medical imaging.

\section{Results}\label{sec2}

\begin{figure}
\centerline{\includegraphics[width=\linewidth]{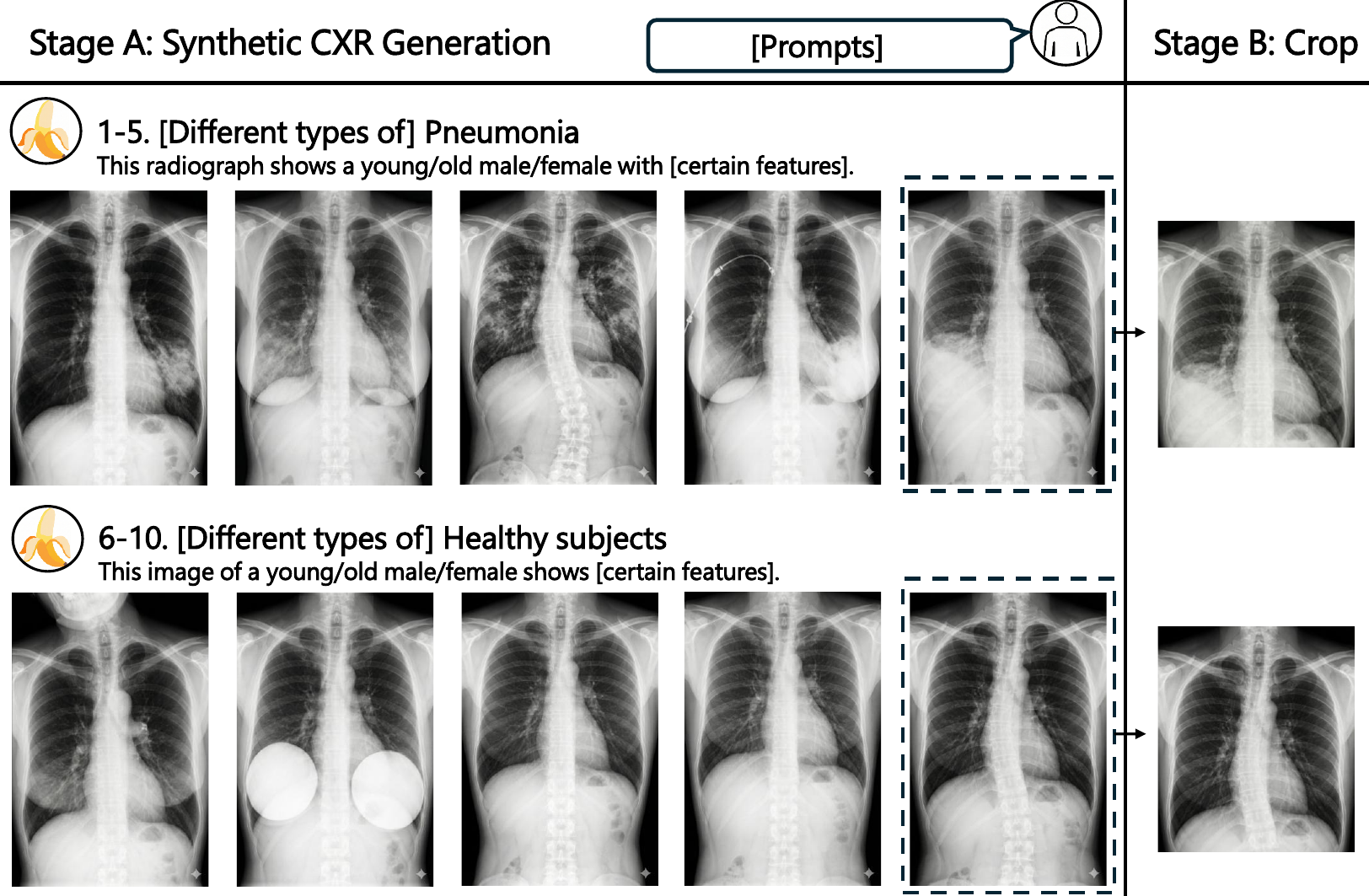}}
\caption{Examples of text-conditional generation of chest radiographs. The top row (1-5) shows synthesized samples of pneumonia, while the bottom row (6-10) depicts healthy subjects. The generated images exhibit diversity across multiple factors including sex, anatomical structures, and medical artifacts. After generation, all images were cropped to remove watermarks and irrelevant regions.}
\label{prompt}
\end{figure}

In this study, we investigated the feasibility of training a pneumonia classifier using synthetic CXRs generated by Nano Banana and evaluated its performance on real-world data. We generated 300 frontal view CXRs with variation in demographics, anthropometrics, imaging positions (posteroanterior and anteroposterior views), and pathological features. Representative examples of healthy and pneumonia cases are shown in Fig.\ref{prompt}. Nano Banana successfully generated CXRs that reflected real-world imaging characteristics, including acquisition-related pose tilt (row 2, column 1), breast prostheses (row 2, column 2), and suboptimal inspiratory effort (row 2, column 4). 
However, the majority of generated images exhibited a field-of-view extending beyond the thoracic region and also included Nano Banana's digital watermark. To align the synthetic data with standard CXR acquisition protocols, we cropped the lower 30\% of each image post-generation. The synthetic dataset was balanced with a 50:50 ratio of pneumonia and healthy subjects, and partitioned into training (n=220), validation (n=30), and test (n=50) sets.

To validate classifier performance, we conducted external validation using two real-world public datasets: the Chest X-Ray dataset (5,856 anteroposterior CXRs; 1,583 healthy, 4,273 pneumonia) \cite{kermany2018identifying} and the 2018 RSNA Pneumonia Detection Challenge dataset (14,863 images; 8,851 normal, 6,012 lung opacity) \cite{shih2019augmenting}. For the RSNA dataset, images labeled as ‘Lung Opacity’ were categorised as pneumonia cases and ‘Normal’ images as healthy controls. Images labeled ‘No Lung Opacity / Not Normal’ were excluded. To ensure robust and holistic external evaluation, we combined all data from a real-world dataset into a single evaluation set rather than using their predefined training and test partitions.

We leveraged a ResNet-50, pre-trained on the ImageNet dataset \cite{deng2009imagenet}, as the base classifier. We then attached a two-layer multi-layer perceptron (MLP) and fine-tuned the network on our curated synthetic dataset for 20 epochs with a fixed learning rate of $5 \times 10^{-5}$. Augmentations were applied to the training images, including random resized cropping, affine transformation, horizontal flipping, and rotation. All images used in this study were resized to 224$\times$224 pixels and normalized using the standard ImageNet mean and standard deviation. Performance was evaluated using Area Under the Receiver Operating Characteristic curve (AUROC) and the Area Under the Precision-Recall curve (AUPR) with 95\% Confidence Interval (CI) calculated via non-parametric bootstrapping with 1,000 iterations. The best-performing model on the validation set, attaining 0.994 AUROC (95\% CI: 0.973 - 1.00) and 0.995 AUPR (95\% CI: 0.984 - 1.00) on the test set, was selected for external validation.

\begin{figure}[]
\centerline{\includegraphics[width=\linewidth]{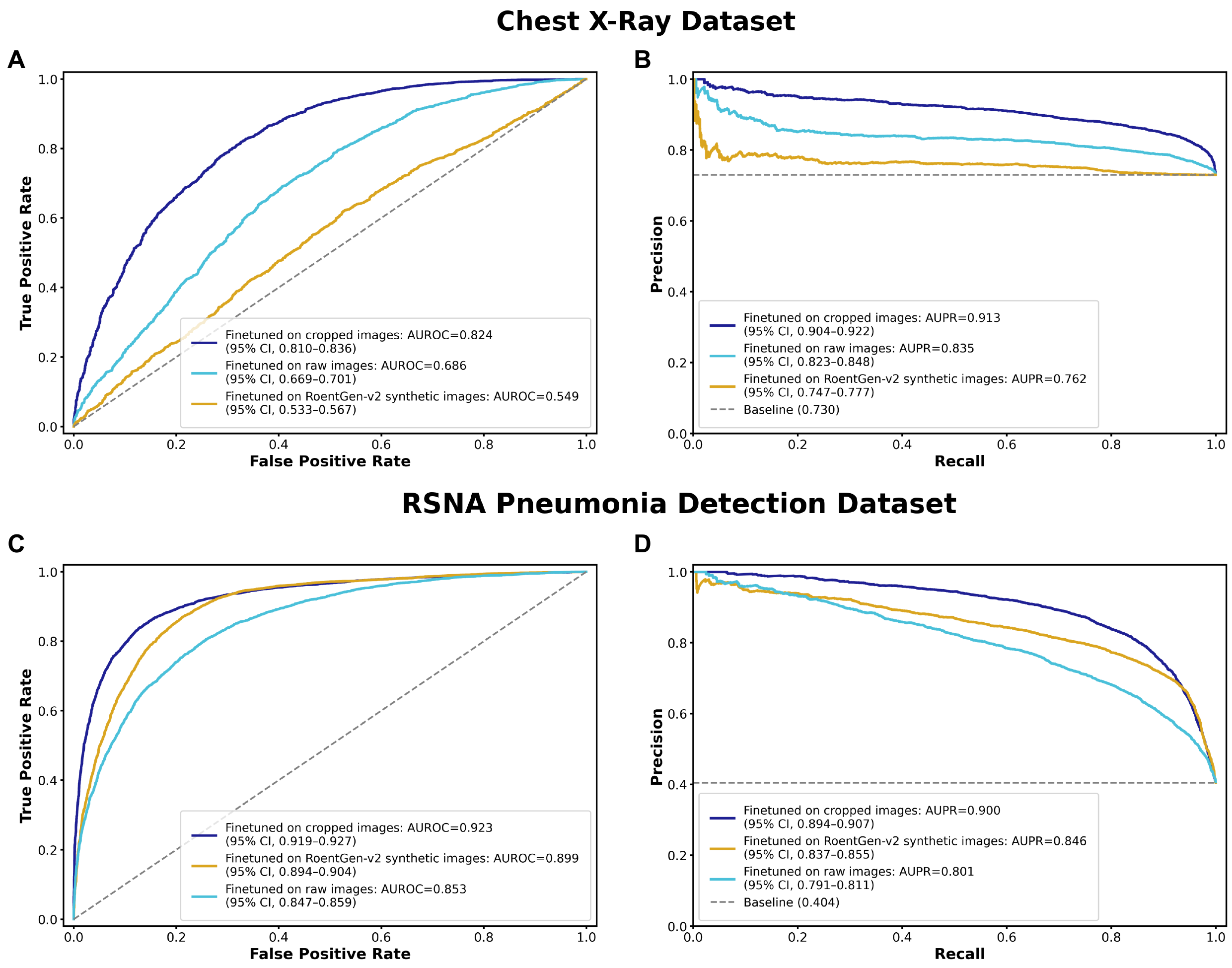}}
\caption{Performance comparison between the models fine-tuned on raw CXRs generated by Nano Banana (light blue) , on cropped Nano Banana CXRs (dark blue), and on RoentGen-v2 generated CXRs (yellow) based on AUROC (A, C) and AUPR (B, D). We conducted evaluation separately on the Chest X-Ray dataset (n=5,856) and the 2018 RSNA Pneumonia Detection Challenge dataset (n=14,863). The baselines in AUPR graphs indicate the proportion of pneumonia cases in the dataset.}
\label{supervised}
\end{figure}

\begin{figure}[]
\centerline{\includegraphics[width=\linewidth]{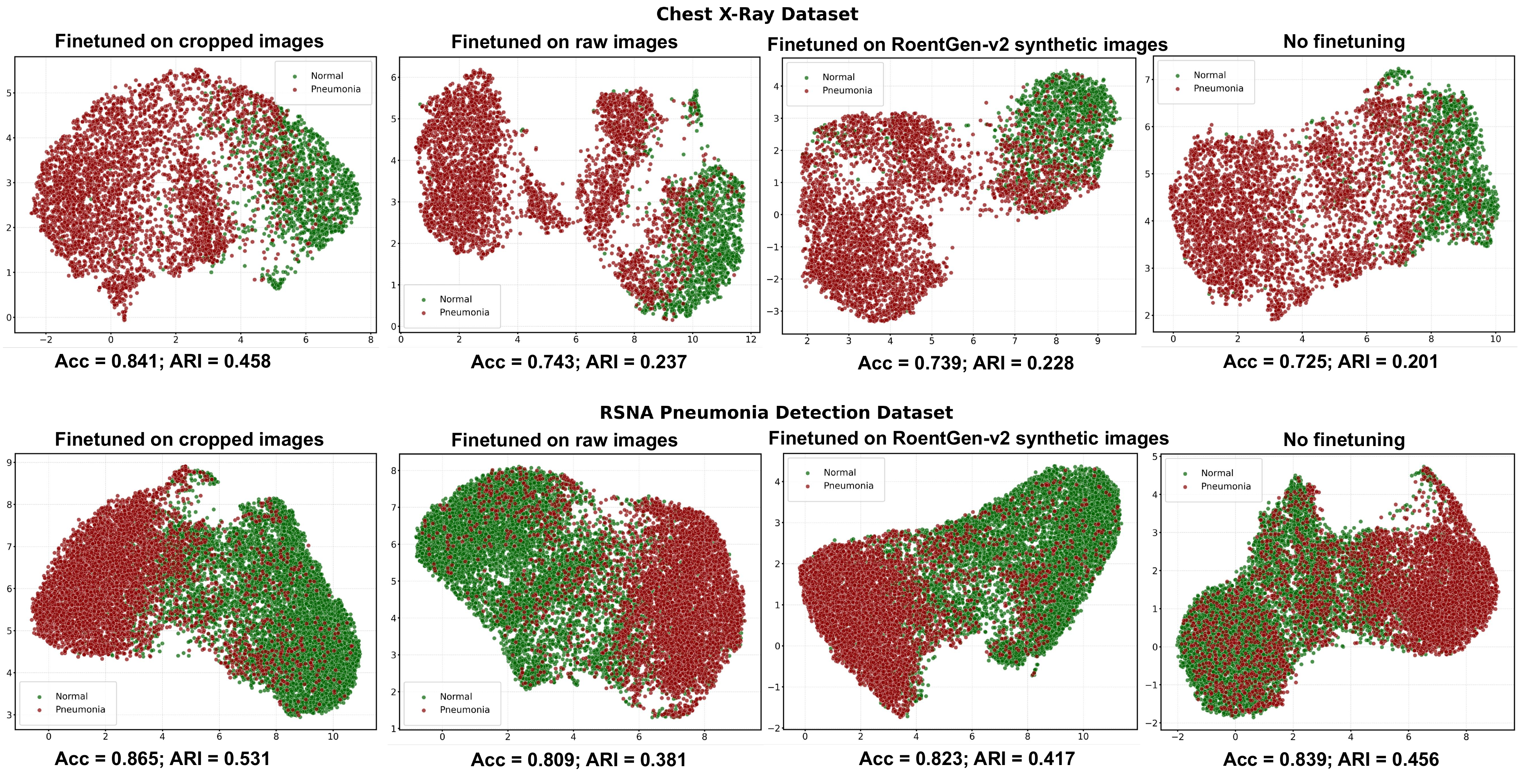}}
\caption{Uniform Manifold Approximation and Projection (UMAP) visualization of features extracted from the classifier fine-tuned on cropped Nano Banana CXRs, on raw synthetic CXRs generated by Nano Banana, on RoentGen-v2 generated synthetic CXRs, and pre-trained on ImageNet (no fine-tuning). We conducted evaluation separately on the Chest X-Ray dataset (n=5,856) and the 2018 RSNA Pneumonia Detection Challenge dataset (n=14,863). For each feature set, we performed K-means clustering and report the resulting classification Accuracy (Acc) and adjusted Rand index (ARI).}
\label{kmeans}
\end{figure}

\begin{figure}[]
\centerline{\includegraphics[width=\linewidth]{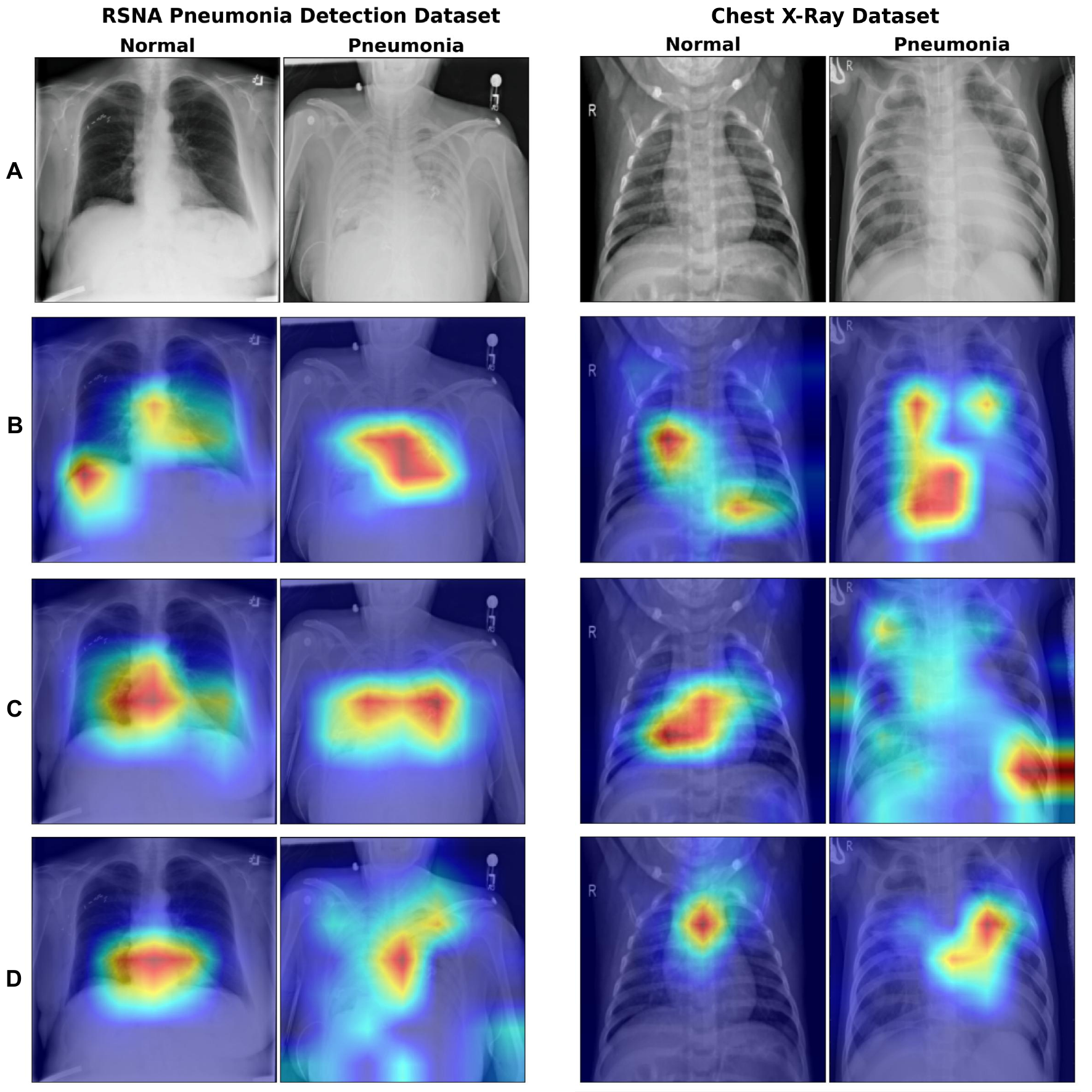}}
\caption{Grad-CAM visualization of the feature map of the last convolutional layer before the classification head. A: original image; B: fine-tuned on cropped CXRs generated by Nano Banana; C: on raw Nana Banana-generated CXRs; and D: on CXRs generated by RoentGen-v2.}
\label{gradcam}
\end{figure}

Fig.~\ref{supervised} shows the results of pneumonia classification on two public real-world datasets. After fine-tuning with cropped synthetic CXRs, the classifer was able to achieve an average AUROC score of 0.923 (95\% CI: 0.919 - 0.927), and an average AUPR score of 0.900 (95\% CI: 0.894 - 0.907) on the RSNA dataset. On the Chest X-Ray dataset, it achieved an average AUROC of 0.824 (95\% CI: 0.810 - 0.836), and an average AUPR of 0.913 (95\% CI: 0.904 - 0.922), demonstrating meaningful utility of Nano Banana-generated CXRs for pneumonia detection.
By contrast, classifiers trained with generated CXRs with extended views and digital watermarks showed poorer performance with a decreased AUROC of 0.853 (95\% CI: 0.847 - 0.859) and an AUPR of 0.801 (95\% CI: 0.791 - 0.811) on the RSNA dataset. On the Chest X-Ray dataset, the AUROC decreased to 0.686 (95\% CI: 0.669 - 0.701) and the AUPR decreased to 0.835 (95\% CI: 0.823 - 0.848). We also curated 300 synthetic CXRs (150 healthy cases and 150 pneumonia cases) from a publicly available dataset generated by RoentGen-v2~\cite{moroianu2025improving}, a state-of-the-art text-to-image diffusion model for CXR generation. We then fine-tuned another ResNet-50 on RoentGen-v2's synthetic CXRs with the same fine-tuning protocol. As shown in Fig.~\ref{supervised}, the classifier fine-tuned with cropped Nano Banana CXRs outperformed the RoentGen-v2-based model on both real-world datasets, whereas the classifier fine-tuned on raw Nano Banana CXRs achieved higher performance on one dataset but lower performance than the RoentGen-v2-based model on another.

We further investigated the effectiveness of the synthetic CXRs, ensuring that the performance was not merely driven by discriminative features already learned from ImageNet. First, we pre-extracted features of images in the two public datasets using the ResNet-50 fine-tuned with Nano Banana's cropped synthetic CXRs, with raw synthetic CXRs generated by Nano Banana, with RoentGen-v2 synthetic CXRs, and the ResNet-50 pre-trained on ImageNet without fine-tuning. We then applied K-means to perform unsupervised feature clustering and obtained the class predictions. As K-means produces hard labels independent of threshold values, we assess the unsupervised classification using Accuracy (Acc). Furthermore, we include the adjusted Rand index (ARI) to evaluate how well the clustering results match the true data distribution. Additionally, we employed UMAP \cite{mcinnes2018umap} to reduce the dimensionality of the extracted features and visualize them in two-dimensional spaces.
As shown in Fig.~\ref{kmeans}, the ResNet-50 fine-tuned on cropped synthetic CXRs generated by Nano Banana produced more discriminative feature representations, showing less overlapping and intertwining between clusters. Furthermore, it achieves higher Acc and ARI scores on both datasets, corroborating the efficacy of the Nano Banana generated synthetic CXRs. By contrast, although the model fine-tuned on raw images and the non-fine-tuned model exhibited different patterns in the UMAP visualizations, the former showed no significant advantage in the quantitative metrics predicted by K-means.

We further used Grad-CAM~\cite{selvaraju2017grad} to visualize heat maps, highlighting the areas where classifiers trained with different synthetic data focus on (Fig.~\ref{gradcam}). For the classifier fine-tuned on cropped Nano Banana CXRs (row B), Grad-CAM indicated dominant attention over clinically salient features including areas of consolidation within pneumonia cases and cardiac and diaphragmatic borders within healthy cases.

\section{Discussion}\label{sec3}
In this study, we demonstrate a classifier fine-tuned with synthetic CXRs generated by Nano Banana can generalize to real-world CXR data with reasonable performance. These preliminary results could suggest that domain-specific models could be developed from merely a few hundred synthetic images produced by large-scale industrial models, greatly mitigating the challenges associated with data acquisition and, hence, improving overall accessibility.

We hypothesize that achieving perfect fidelity of synthetic medical data may not be necessary for effective model training. For instance, synthetic CXRs that are capable of capturing disease-relevant pathological features could potentially teach classifiers to learn generalizable disease signatures while avoiding shortcut learning~\cite{geirhos2020shortcut}. Grad-CAM analysis supports this possibility, as classifiers trained on synthetic CXRs predominantly focused on lung fields, regions of consolidation in pneumonia cases, and key cardiothoracic structures in healthy cases, broadly aligning with clinically relevant areas of inspection for physicians.

While results are promising, several practical and methodological limitations emerged. Firstly, prompt engineering proved insufficient for reliably controlling imaging posture and angle variation across generated CXR images. Secondly, we found substantial performance degradation when using uncropped synthetic images. Accordingly, at this current stage, careful prompt design and appropriate post-processing of the generated data appears to be required to enhance the utility of the synthetic data. Thirdly, this study only investigated the use of CXRs generated by Nano Banana in recognizing pneumonia. Generalization to wider medical domains, modalities and tasks were not explored. Hence, the general effectiveness of synthetic medical data generated by Nano Banana will require further exploration.

While synthetic data can democratize AI development, this must be contextualized within the unresolved ethical, regulatory, and legal landscape. Regulatory pathways for AI medical devices mandate validation against real-world clinical datasets with synthetic data alone insufficient for FDA or CE marking. Digital watermarks also raise questions surrounding intellectual property and liability and accountability frameworks remain undefined with current healthcare law and medical malpractice frameworks not addressing liability for AI systems trained on synthetic data.

\section{Conclusion}\label{sec4}

In this study, we investigated the utility of synthetic CXRs generated by Nano Banana for developing an AI classifier for recognizing pneumonia. While the external validation results are promising, further exploration is required across wider medical domains, modalities and tasks to demonstrate the value of synthetic medical data generated by industrial models. Beyond the technical challenges, complementary policies and guidelines for transparent and responsible use of synthetic medical data delivered through Internet-based interfaces should be developed in tandem.

\section{Appendix}

We used the prompt \texttt{Generate 10 Chest X-ray imaging data consisting of images representing different healthy/pneumonia patients, generated individually as separate downloadable files so I can download them one by one. The views you generate need to maintain a consistent format, meaning the overall image is portrait-oriented, and the images must show variations in gender, height, weight (fat and thin), and age, as well as human posture and body stance during imaging (such as some bodies or heads tilted left or right, rotation, arm orientation, etc.), along with differences in lung texture to ensure clinical authenticity and individual diversity. Note that each picture must be on a separate canvas, meaning you need to generate 10 images, all in portrait orientation with height greater than width, and the view focused on the thoracic cavity.} for generating synthetic CXRs from Nano Banana.

\backmatter


\end{document}